\documentclass{article}

\PassOptionsToPackage{square,comma,numbers,sort&compress}{natbib}

\usepackage[final]{include/arXiv}

\usepackage[utf8]{inputenc} \usepackage[T1]{fontenc}    \usepackage{hyperref}       \usepackage{url}            \usepackage{booktabs}       \usepackage{amsfonts}       \usepackage{nicefrac}       \usepackage{microtype}      
\usepackage{caption}
\usepackage{subcaption}

\usepackage{algorithm}
\usepackage{algorithmicx}
\usepackage[noend]{algpseudocode}
\usepackage{changepage}

\algrenewcommand\algorithmicindent{1.0em}

\usepackage{graphicx}
\usepackage{comment}
\usepackage{amsmath}
\usepackage{amssymb}
\usepackage{siunitx}
\usepackage{wrapfig}
\usepackage{todonotes}
\usepackage{enumitem}
\usepackage{siunitx}
\usepackage{dsfont}
\usepackage{bbm}

\newcommand{\E}{\mathbb{E}}

\newcommand{\be}{\begin{equation}}
\newcommand{\ee}{\end{equation}}
\newcommand{\bea}{\begin{eqnarray}}
\newcommand{\eea}{\end{eqnarray}}
\newcommand{\beaa}{\begin{eqnarray*}}
\newcommand{\eeaa}{\end{eqnarray*}}

\DeclareMathAlphabet{\mathpzc}{OT1}{pzc}{m}{n}

\usepackage{xcolor,colortbl}
\usepackage[english]{babel}
\usepackage{cleveref}

\title{Multi-Objective Deep Reinforcement Learning}

\author{
  Hossam Mossalam, Yannis M. Assael, Diederik M. Roijers, Shimon Whiteson\\
  Department of Computer Science\\
  University of Oxford\\
  Oxford, United Kingdom \\
  \texttt{\{ms15ham, yannis.assael, diederik.roijers, shimon.whiteson\}@cs.ox.ac.uk}
}

\begin{document}

\maketitle

\begin{abstract}
	We propose \textit{Deep Optimistic Linear Support Learning (DOL)} to solve high-dimensional \textit{multi-objective} decision problems where the relative importances of the objectives are not known a priori.
Using features from the high-dimensional inputs, DOL computes the convex coverage set  containing all potential optimal solutions of the convex combinations of the objectives.
To our knowledge, this is the first time that deep reinforcement learning has succeeded in learning multi-objective policies.
In addition, we provide a testbed with two experiments to be used as a benchmark for deep multi-objective reinforcement learning.

\end{abstract}

\section{Introduction}

In recent years, advances in deep learning have been instrumental in solving a number of challenging reinforcement learning~(RL) problems, including high-dimensional robot control  \cite{levine2015end,assael2015data,Watter:2015}, visual attention \cite{Ba:2015}, solving riddles~\cite{foerster2016learning}, the \emph{Atari learning environment} (ALE)~\cite{Guo:2014,Mnih:2015,stadie2015incentivizing,Wang:2015duel,Schaul:2015,van2015deep,oh2015action,Bellemare2015Persistent,Nair:2015} and Go~\cite{Maddison:2015,silver2016mastering}.

While the aforementioned approaches have focused on single-objective settings, many real-world problems have \emph{multiple} possibly conflicting \emph{objectives}.
For example, an agent that may want to maximise the performance of a web application server, while minimising its power consumption~\cite{Tesauro}.
Such problems can be modelled as \emph{multi-objective Markov decision processes~(MOMDPs)}, and solved with \emph{multi-objective reinforcement learning~(MORL)}~\cite{jair13}.
Because it is typically not clear how to evaluate available trade-offs between different objectives a priori, there is no single optimal policy.  Hence, it is desirable to produce a \emph{coverage set~(CS)} which contains at least one optimal policy (and associated value vector) for each possible utility function that a user might have.

So far, deep learning methods for Markov decision processes (MDPs) have not been extended to MOMDPs. One  reason is that it is not clear how neural networks can account for unknown preferences and the resulting sets of value vectors. In this paper, we circumvent this issue by taking an \emph{outer loop approach}~\cite{roijers2015computing} to multi-objective reinforcement learning, i.e., we aim to learn an approximate coverage set of policies, each represented by a neural network, by evaluating a sequence of scalarised single-objective problems.
In order to enable the use of deep Q-Networks~\cite{Mnih:2015} for learning in MOMDPs, we build off the state-of-the-art \emph{optimistic linear support (OLS)}  framework~\cite{roijersPhD,roijers2015computing}.
OLS is a generic \emph{outer loop method} for solving multi-objective decision problems, i.e., it repeatedly calls a single-objective solver as a subroutine. OLS terminates after a finite number of calls to that subroutine and produces an approximate CS. In principle any single-objective solver can be used, as long as it is OLS-compliant, i.e., produces policy value vectors rather than scalar values. Making a single-objective solver OLS-compliant typically requires little effort.

We present three new deep multi-objective RL algorithms. First, we investigate how the learning setting effects OLS, and how deep RL can be made \emph{OLS-compliant}.
Using an OLS-compliant neural network combined with the OLS framework results in \emph{Deep OLS Learning} (DOL). Our empirical evaluation shows that DOL can tackle multi-objective problems with much larger inputs than classical multi-objective RL algorithms.
We improve upon DOL by leveraging the fact that the OLS framework solves a series of single-objective problems that become increasingly similar as the series progresses~\cite{roijers2015point}, which results in increasingly similar optimal value vectors.
Deep Q-networks produce latent embeddings of the features of a problem w.r.t.\ the function value. Hence, we hypothesise that we can \emph{reuse} parts of the network used to solve the previous single-objective problem, in order to speed up learning on the next one. This results in two new algorithms that we call \emph{Deep OLS Learning with Full Reuse} (DOL-FR), which reuses all parameter values of neural networks, and \emph{Deep OLS Learning with Partial Reuse} (DOL-PR) which reuses all parameter values of neural networks, except those for the last layer of the network. We show empirically that reusing only part of the network (DOL-PR) is more effective than reusing the entire network (DOL-FR) and drastically improves the performance compared to DOL without reuse.

\section{Background}

In a single-objective RL setting \cite{sutton1998introduction}, an agent observes the current state $s_t \in \mathcal{S}$ at each discrete time step $t$, chooses an action $a_t \in \mathcal{A}$ according to a potentially stochastic policy $\pi$, observes a reward signal $R(s_{t}, a_{t}) = r_t \in \mathcal{R}$, and transitions to a new state $s_{t+1}$.
Its objective is to maximise an expectation over the discounted return, $R_t =  r_t + \gamma r_{t+1} + \gamma^2 r_{t+2} + \cdots$, where $r_t$ is the reward received at time $t$ and $\gamma \in [0,1]$ is a discount factor.

\textbf{Markov Decision Process (MDP).} Such sequential decision problems are commonly modelled as a finite \emph{single-objective Markov decision process (MDP)}, a tuple of $\langle \mathcal{S}, \mathcal{A}, R, T,  \gamma \rangle$.
The $Q$-function of a policy $\pi$ is $Q^{\pi}(s,a) = \E \left[  R_t | s_t = s,a_t = a \right]$. The optimal action-value function $Q^{*}(s,a) = \max_{\pi} Q^{\pi}(s,a)$ obeys the Bellman optimality equation:
\begin{equation}
    Q^{*}(s,a) = \E_{s'} \left[ R(s,a) +  \gamma  \max_{a' } Q^{*}(s',a')~|~s,a \right].
\end{equation}

\textbf{Deep Q-Networks (DQN).} Deep $Q$-learning \cite{Mnih:2015} uses neural networks parameterised by $\theta$ to represent $Q(s,a;\theta)$. DQNs are optimised by minimising:
\begin{equation}
\mathcal{L}_i(\theta_i) = \E_{s,a,r,s'} \left[( y_i^{DQN} - Q(s,a; \theta_{i}))^{2}\right],
\label{eq:loss}
\end{equation}
at each iteration~$i$, with target $y_i^{DQN} =  r + \gamma \max_{a'} Q(s',a';\theta_{i}^{-})$. Here, $\theta_{i}^{-}$ are the parameters of a target network that is frozen for a number of iterations while updating the online network $Q(s,a; \theta_{i})$ by gradient descent. The action $a$ is chosen from $Q(s,a; \theta_{i})$ by an \emph{action selector}, which typically implements an $\epsilon$-greedy policy that selects the action that maximises the Q-value with a probability of $1-\epsilon$ and chooses randomly with a probability of $\epsilon$. DQN uses \emph{experience replay}~\cite{lin1993reinforcement}: during learning, the agent builds a dataset of episodic experiences and is then trained by sampling mini-batches of experiences.
Experience replay is used in~\cite{Mnih:2015} to reduce variance by breaking correlation among the samples, whilst, it enables re-use of past experiences for learning. 

\begin{wrapfigure}{r}{.4\textwidth}
   \vspace{-7mm}
    \begin{center}
        \includegraphics[width=.39\textwidth]{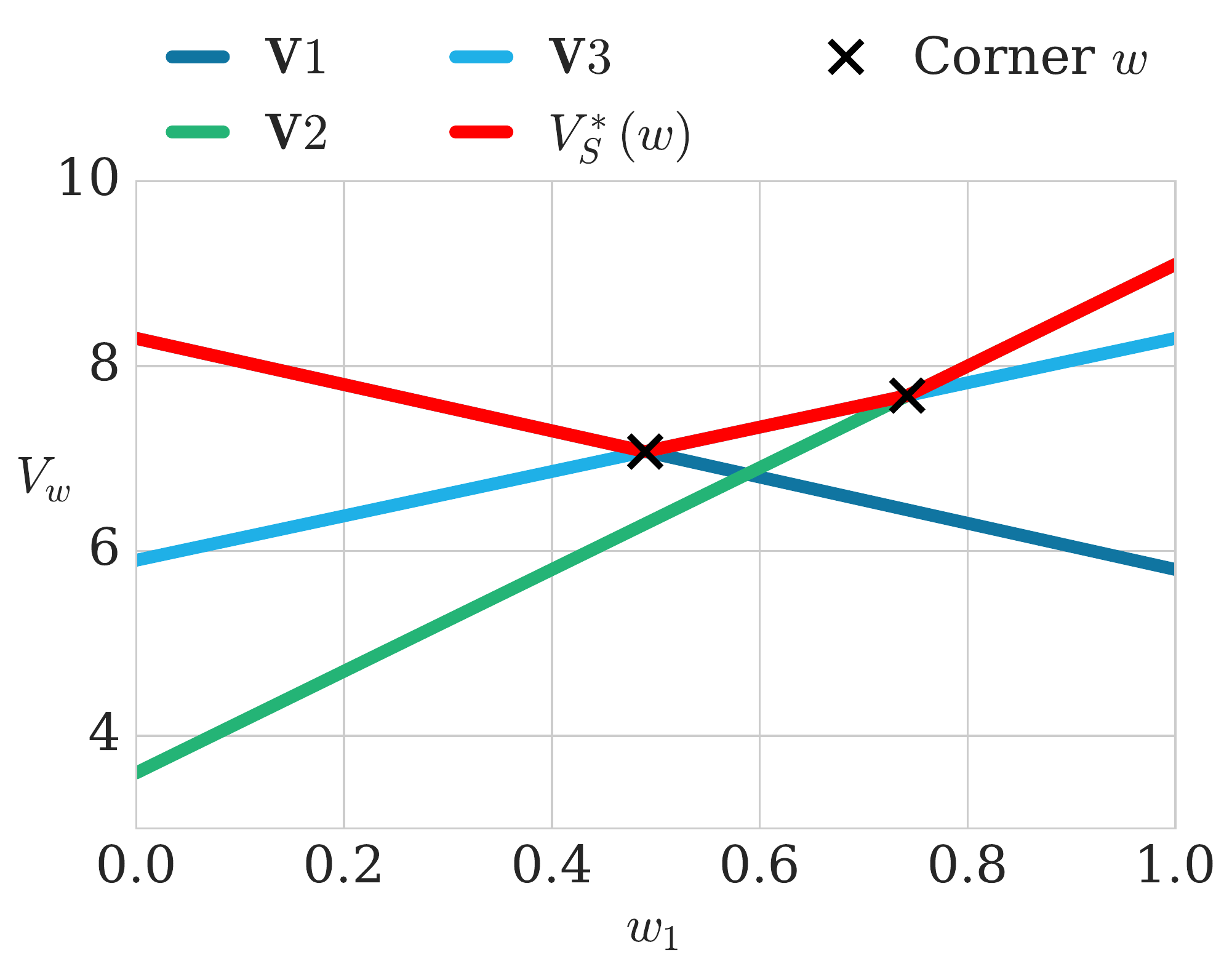}
    \end{center}
    \vspace{-3.5mm}
    \caption{\small The two corner weights of a $V_{S}^*({\bf w})$ with $S$ containing three value vectors for a 2-objective MOMDP.}
     \vspace{-2mm}
    \label{fig:corner-weights}
\end{wrapfigure}
\textbf{Multi-Objective MDPs (MOMDP).} An MOMDP, is an MDP in which the reward function ${\bf R}(s_{t}, a_{t}) = {\bf r}_t \in \mathcal{R}^n$ describes a vector of $n$ rewards, one for each objective~\cite{jair13}. We use bold variables to denote vectors.
The solution to an MOMDP is a set of policies called a \emph{coverage set}, that contains at least one optimal policy for each possible preference, i.e., utility or scalarisation function, $f$, that a user might have. This scalarisation function maps each possible policy value vector, ${\bf V}^\pi$ onto a scalar value. In this paper, we focus on the highly prevalent case where the scalarisation function, is linear, i.e., $f({\bf V}^\pi, {\bf w}) = {\bf w} \cdot {\bf V}^\pi$, where $\bf w$ is a vector that determines the relative importance of the objectives, such that $f({\bf V}^\pi, {\bf w})$ is a convex combination of the objectives. The corresponding coverage set is called the \emph{convex coverage set} (CCS) \cite{jair13}.

\textbf{Optimistic Linear Support (OLS). }
OLS takes an \emph{outer loop approach} in which the CCS is incrementally constructed by solving a series of scalarised, i.e., single-objective, MDPs for different linear scalarisation vectors $\bf w$. This enables the use of DQNs as a single-objective MDP solver. In each iteration, OLS finds one policy by solving a scalarised MDP, and its value vector ${\bf V}^\pi$ is added to an intermediate approximate coverage set, $S$.

Unlike other outer loop methods, OLS uses the concept of \emph{corner weights} to pick the weights to use for creating scalarised instances and the concept of estimated improvement to prioritise those corner weights. To define corner weights, we first define the scalarised value function  $V_{S}^*({\bf w}) = \max_{V^\pi \in S} \,{\bf w} \cdot {\bf V}^\pi $, as a function of the linear scalarisation vector $\bf w$, for a set of value vectors $S$.
$V_{S}^*({\bf w})$ for an $S$ containing three value vectors is depicted in Figure \ref{fig:corner-weights}. $V_{S}^*({\bf w})$ forms a piecewise linear and convex function that comprise the upper surface of the scalarised values of each value vector.
The corner weights are the weights at the corners of the convex upper surface \cite{Cheng}, marked with crosses in the figure. OLS always selects the corner weight $\bf w$ that maximises an optimistic upper bound on the difference between $V_S^*({\bf w})$ and the optimal scalarised value function, i.e.,  $V_{CCS}^*({\bf w}) - V_{S}^*({\bf w})$, and solves the single-objective MDP scalarized by the selected $\bf w$.

In the planning setting for which OLS was devised, such an upper bound can typically be computed using upper bounds on the error with respect to the optimal value of the scalarised policy values at each previous $\bf w$ in the series, in combination with linear programs. The error bounds at the previous $\bf w$ stem from the approximation quality of the single-objective planning methods that OLS uses.
However, in reinforcement learning, the true $CCS$ is fundamentally unknown and no upper bounds can be given on the approximation quality of deep $Q$-learning. Therefore, we use $V_{\overline{CCS}}^*({\bf w}) - V_{S}^*({\bf w})$ as a heuristic to determine the priority, where  $V^*_{\overline{CCS}}(\mathbf{w})$ is defined as maximal attainable scalarised value if we assume that the values found for previous~$\bf w$ in the series were optimal for those~$\bf w$.

\section{Methodology}

In this section, we propose our algorithms for MORL that employ deep Q-learning. Firstly, we propose our basic \emph{deep OLS learning (DOL)} algorithm; we build off the OLS framework for multi-objective learning and integrate DQN. Then, we improve on this algorithm by introducing \emph{Deep OLS Learning with Partial~(DOL-PR)} and \emph{Full Reuse~(DOL-FR)}. DOL, DOL-PR, and DOL-FR make use of a single-objective subroutine, which is defined together with DOL in Section~\ref{sec:scalDQN}.

\subsection{Deep OLS Learning (DOL)}\label{sec:scalDQN}

There are two requirements to make use of the OLS framework. We first need a scalarized, i.e., single-objective learning algorithm that is OLS compliant. OLS compliance entails that rather than learning a single value per $Q(s,a)$, we need a vector-valued Q-value ${\bf Q}(s,a)$.
The estimates of ${\bf Q}(s,a)$ need to be accurate enough to determine the next corner weight in the series of linear scalarisation weights, $\bf w$, that OLS is going to generate.
To satisfy those requirements we adjust our neural network architectures to output a matrix of $|\mathcal{A}| \times n$ (where $n$ is the number of objectives) instead of just $|\mathcal{A}|$, and we train for an extended number of episodes.

We define \emph{scalarised deep Q-learning}, which uses this network architecture, and optimises the parameters to maximise the inner product of $\bf w$ and the ${\bf Q}$-values for a given $\bf w$ instead of the scalar $Q$-values as in standard deep Q-learning.
Using scalarised deep Q-learning as a subroutine in OLS results in our first algorithm: \emph{deep OLS learning (DOL)}.

\subsection{Deep OLS Learning with Full (DOL-FR) and Partial Reuse (DOL-PR)}

While DOL can already tackle very large MOMDPs, re-learning the parameters for the entire network when we move to the next $\bf w$ in the sequence is rather inefficient.
Fortunately, we can exploit the following observation: the optimal value vectors (and thus optimal policies) for a scalarised MOMDP with a $\bf w$ and a $\bf w'$ that are close together, are typically close as well~\cite{roijers2015point}.
Because deep Q-networks learn to extract the features of a problem that are relevant to the rewards of an MOMDP, we can speed up computation by reusing the neural network parameters that were trained earlier in the sequence.

In Algorithm~\ref{alg:dol}, we present an umbrella version of three novel algorithms, which we denote $\tt DOL$. The different algorithms are obtained by setting the $\tt reuse$ parameter (i.e., the type of reuse) to one of three values: DOL (without reuse) is obtained by setting  $\tt reuse$ to `none', \emph{DOL with full reuse (DOL-FR)} is obtained by setting  $\tt reuse$ to `full', and  \emph{DOL with partial reuse (DOL-PR)}  is obtained by setting  $\tt reuse$ to `partial'.

\begin{algorithm}[tb]
   \caption{Deep OLS Learning (with different types of reuse)}
   \label{alg:dol}
   \hskip -2em
\begin{algorithmic}[1]
\Function{DOL}{$m,\tau,{\tt template}, {\tt reuse}$} \label{ln:functioncall}
	\State  $\triangleright$ Where, $m$ -- the (MOMDP) environment, $\tau$ -- improvement threshold,
	\State  $\triangleright$ ${\tt template}$ -- specification of DQN architecture, $\tt reuse$ -- the type of reuse
	\State  $S =$ empty partial CSS \label{ln:sinit}
	\State  $W =$ empty list of explored corner weights \label{ln:winit}
	\State  $Q =$ priority queue initialised with the extrema weights simplex with infinite priority \label{ln:qinit}
	\State  ${\tt DQN\_Models} =$ empty table of DQNs, indexed by the weight, $\bf w$, for which it was learnt \label{ln:dinit}
	\While{$Q$ is not empty $\wedge {\tt it} \le {\tt max\_it}$}
		\State ${\bf w} = Q.\text{pop}()$\label{ln:pop}
		\If{${\tt reuse} =$ `none' $\vee ~{\tt DQN\_Models}$ is empty }
			\State ${\tt model} =$ a randomly initialised DQN, from a pre-specified architecture {\tt template}\label{ln:randommodel}
		\Else
			\State ${\tt model} = \text{copyNearestModel}({\bf w}, \text{DQN\_Models})$\label{ln:reuse} 
			\If{${\tt reuse} =$ `partial'}
				 reinitialise the last layer of ${\tt model}$ with random weights \label{ln:partial}
			\EndIf
		\EndIf
		\State ${\bf V}, {\tt new\_model} = \text{scalarisedDeepQLearning}(m, {\bf w}, {\tt model} )$ \label{ln:call}
		\State $W =  W \cup {\bf w}$ \label{ln:wadd}
		\If{$\displaystyle (\exists {\bf w}')~~ {\bf w}' \!\!\cdot\!{\bf V} > \max_{{\bf U} \in S} {\bf w}'\!\!\cdot\!{\bf U}$}
			\State $W_{del}\ =  W_{del} \cup$ corner weights made obsolete by $\bf V$ from $Q$\label{ln:accept1}
			\State $W_{del}\ =  W_{del} \cup \{{\bf w}\}$\label{ln:accept2}
			\State Remove $W_{del}$ from $Q$
			\State Remove vectors from $S$ that are no longer optimal for any $\bf w$ after adding $\bf V$
			\State $W_{{\bf V}}$ = newCornerWeights(S, $\bf V$)
			\State $S =  S  \cup  \{{\bf V}\}$
			\State $\text{DQN\_Models}[{\bf w}] =  {\tt new\_model}$
			\For{{\bf each} ${\bf w}' \in W_{\bf V}$ \label{ln:acceptenough}}
				\If{estimateImprovement(${\bf w}', W, S$) > $\tau$}
					\State $Q.\text{add}({\bf w}')$
				\EndIf
			\EndFor
			\label{ln:acceptend}
		\EndIf
    \State ${\tt it}+\!\!+$
	\EndWhile
	\State \Return $S$, ${\tt DQN\_Models}$
	\EndFunction
\end{algorithmic}
\end{algorithm}

DOL-FR applies full deep Q-network reuse; we start learning for a new scalarisation weight $\bf w'$, using the complete network we optimised for the previous $\bf w$ that is closest to $\bf w'$ in the sequence of scalarisation weights that OLS generated so far. DOL-PR applies  partial deep Q-network reuse; we take the same network as for full reuse, but we reinitialise the last layer of the network randomly, in order to escape local optima. DOL (without reuse) does no reuse whatsoever, i.e., all network parameters are initialised randomly at the start of each iteration.

$\tt DOL$ keeps track of the partial CCS, $S$, to which at most one value vector will be added at each iteration (line~\ref{ln:sinit}). To find these vectors, scalarised deep Q-learning (Section~\ref{sec:scalDQN}) is run for different corner weights. The corner weights that are not yet explored are kept in a priority queue, $Q$, and after they have been explored, are stored in a list $W$ (line~\ref{ln:winit} and~\ref{ln:qinit}).  $Q$ is initialised with the extrema weights and keeps track of the scalarisation weights ordered by estimated improvement. In order to reuse the learnt parameters in DOL-PR/FR, $\tt DOL$ keeps track of them along with the corner weight $\bf w$ for which they were found in $\tt DQN\_Models$.

Following the OLS framework, at each iteration of $\tt DOL$, the weight with the highest improvement is popped (line~\ref{ln:pop}). After selecting $\bf w$, $\tt DOL$ now reuses the DQNs it learnt in previous iterations (depending on the parameter $\tt reuse$). The function $\text{copyNearestModel}$ finds the network learnt for the closest weight to the current corner weight on line~\ref{ln:reuse}. In the case of full reuse (${\tt reuse}=`full'$), all parameter values are copied. In the case of partial reuse (${\tt reuse}=`partial'$), the last layer is reinitialised with random parameter values (line~\ref{ln:partial}), and in the case of no reuse (${\tt reuse}=`none'$) all the network parameters are reset (line~\ref{ln:randommodel}).

Following the different types of reuse, scalarised deep Q-learning, as described in Section~\ref{sec:scalDQN} is invoked for the $\bf w$ popped off of $Q$ on line~\ref{ln:pop}.
Scalarised deep Q-learning returns a value vector, $\bf V$, corresponding to the learnt policy represented by a DQN, which is also returned (line~\ref{ln:call}). The current corner weight is added to the list of explored weights (line~\ref{ln:wadd}), which is used to determine the priorities for subsequently discovered corner weights in the current and future iterations.
If there is a weight vector $\bf w$ in the weight simplex for which the scalarised value is higher than for any of the vectors in $S$, the value vector is added to $S$, and new corner weights are determined and stored (lines~\ref{ln:accept1}-\ref{ln:acceptend}). The DQN that corresponds to $\bf V$ is stored in ${\tt DQN\_models}[{\bf w}]$.  If $\bf V$ does not improve upon $S$ for any $\bf w$, it is discarded.

Extending $S$ with $\bf V$ leads to new corner weights. These new corner weights and their estimated improvement are calculated using the $\tt newCornerWeights$ and $\tt estimateImprovement$ methods of OLS~\cite{roijersPhD}. The new corner weights are added to $Q$ if their improvement value is greater than the threshold $\tau$ (lines~\ref{ln:acceptenough}-\ref{ln:acceptend}). Also, corner weights in $Q$ which are made obsolete (i.e. are no longer on the convex upper surface) by the new value vector are removed (line~\ref{ln:accept1}-\ref{ln:accept2}). This is repeated until there are no more corner weights in $Q$, at which point $\tt DOL$ terminates.

\section{Experimental Evaluation}
\label{sec:exp}

In this section, we evaluate the performance of DOL and DOL-PR/FR. We make use of two multi-objective reinforcement learning problems called \emph{mountain car (MC)} and \emph{deep sea treasure (DST)} . We first show, how DOL and DOL-PR/FR are able to learn the correct CSS, using direct access to the state $s_t$ of the problems. Then, we explore the scalability of our proposed methods, and evaluate the performance of weight reuse, we create an image version of the DST problem, in which we use a bitmap as input for scalarised deep Q-learning.

\subsection{Setup}

For both the raw and image problems we follow the DQN setup of~\cite{Mnih:2015}, employing experience replay and a target network to stabilise learning. We use an $\epsilon$-greedy exploration policy with $\epsilon$ annealing from $\epsilon=1$ to $0.05$, for the first $2000$ and $3000$ episodes, respectively, and learning continues for an equal number of episodes. The discount factor is $\gamma = 0.97$, and the target network is reset every $100$ episodes. To stabilise learning, we execute parallel episodes in batches of $32$. The parameters are optimised using Adam and a learning rate of $10^{-3}$. In each experiment we average over $5$ runs.

\begin{wrapfigure}{r}{.33\textwidth}
    \centering
    \vspace{-3mm}
    \includegraphics[width=1\linewidth]{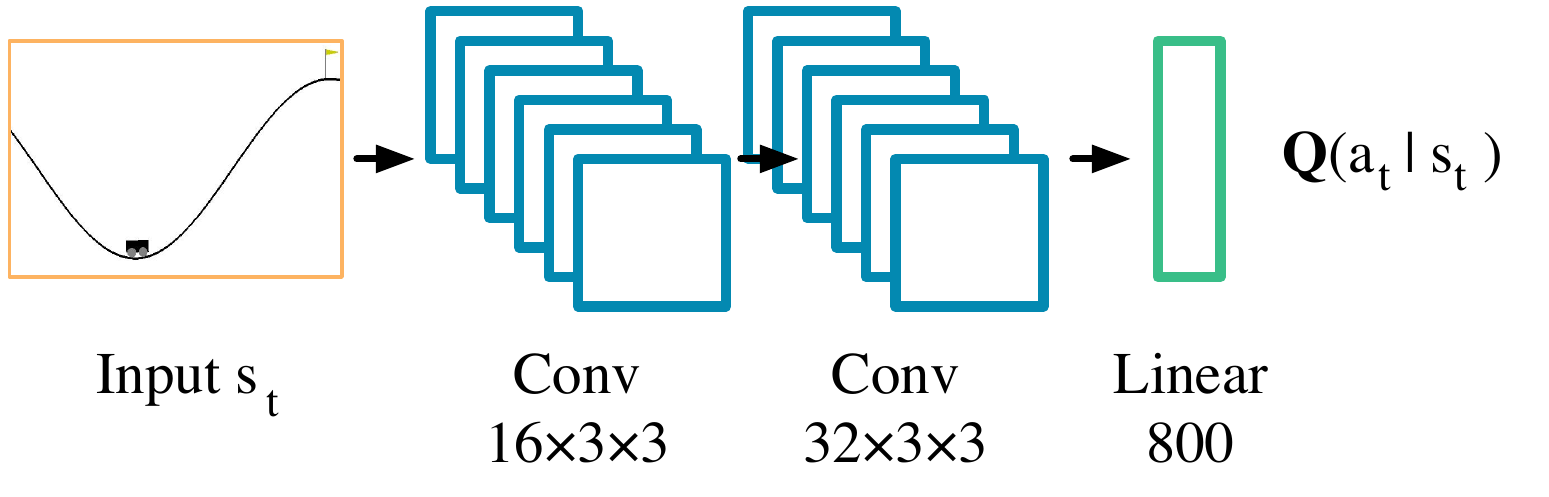}
    \caption{DST architecture.}	
\end{wrapfigure}
For the raw state model we used an MLP architecture with 1 hidden layer of $100$ neurons, and rectified linear unit activations. To process the $3\times11\times10$ image inputs of Deep Sea we employed two convolutional layers of $16 \times 3 \times 3$ and $32 \times 3 \times 3$ and a fully connected layer on top. Finally, to facilitate future research we publish the source-code to replicate our experiments~\footnote{\url{https://github.com/hossam-mossalam/multi-objective-deep-rl}}.

\subsection{Multi-Objective Mountain Car}

In order to show that DOL, DOL-FR, and DOL-PR can learn a CCS, we first test on the \emph{multi-objective mountain car problem (MC)}. MC is a variant of the famous mountain car problem introduced in~\cite{sutton1998introduction}. In single-objective mountain car problem, the agent controls a car located in a valley between two hills and it tries to get the car to reach the top of the hill on the right side. The car has a limited engine power, thus, the agent needs to oscillate the car between both hills until the car has gathered enough inertia that would let it reach the goal. 
\begin{wrapfigure}{r}{.33\textwidth}
	\vspace{-6mm}
	\centering
    \includegraphics[width=1\linewidth]{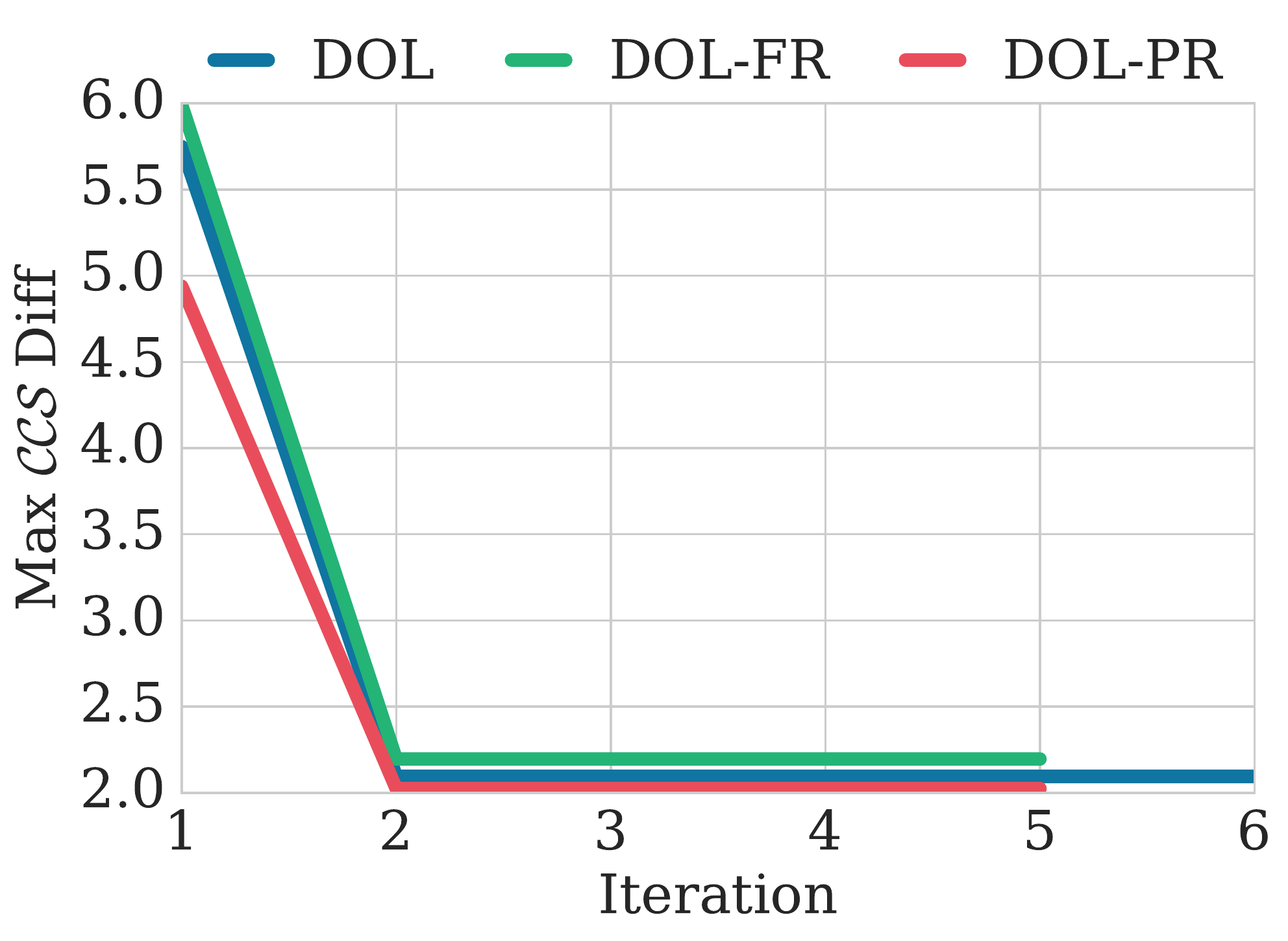}
	\vspace{-6mm}
	\caption{MC raw version mean CSS error.}
	\vspace{-6mm}
    \label{fig:mountaincar-map-max-ccs-err}
\end{wrapfigure}
The reward in the single-objective variant is $-1$ for all time steps and $0$ when the goal is reached. Our multi-objective variant adds another reward which is the fuel consumption for each time step, which is proportional to the force exerted by the car. In MC, there are only 2 value vectors in the CCS, and is thus a small problem. 

\textbf{Raw version.} We evaluate our proposed methods within the MC environment with the agent having direct access to the $s_t$. We employ the same neural network architecture as for DST. However, for MC, we used the CCS obtained by q-table algorithm as the true CCS which was then used for Max CCS error calculations as the true CCS.
As it can be seen in \Cref{fig:mountaincar-map-max-ccs-err}, the three algorithms achieve very similar results on the MC problem with DOL-PR achieving the least error. 
The algorithms learn a good approximation to the CCS in 2 iterations. After that, they continue making tiny improvements to these vectors that are not visible on the graph. The different algorithms behave equally well, which is due to the fact that for the extrema of the weight space, i.e., the first two iterations,  the optimal policies are very different, and reuse does not contribute significantly. 

\vspace{-2mm}
\subsection{Deep Sea Treasure}
\begin{wrapfigure}{r}{.275\textwidth}
	\vspace{-4mm}
	\centering
	\includegraphics[width=1\linewidth]{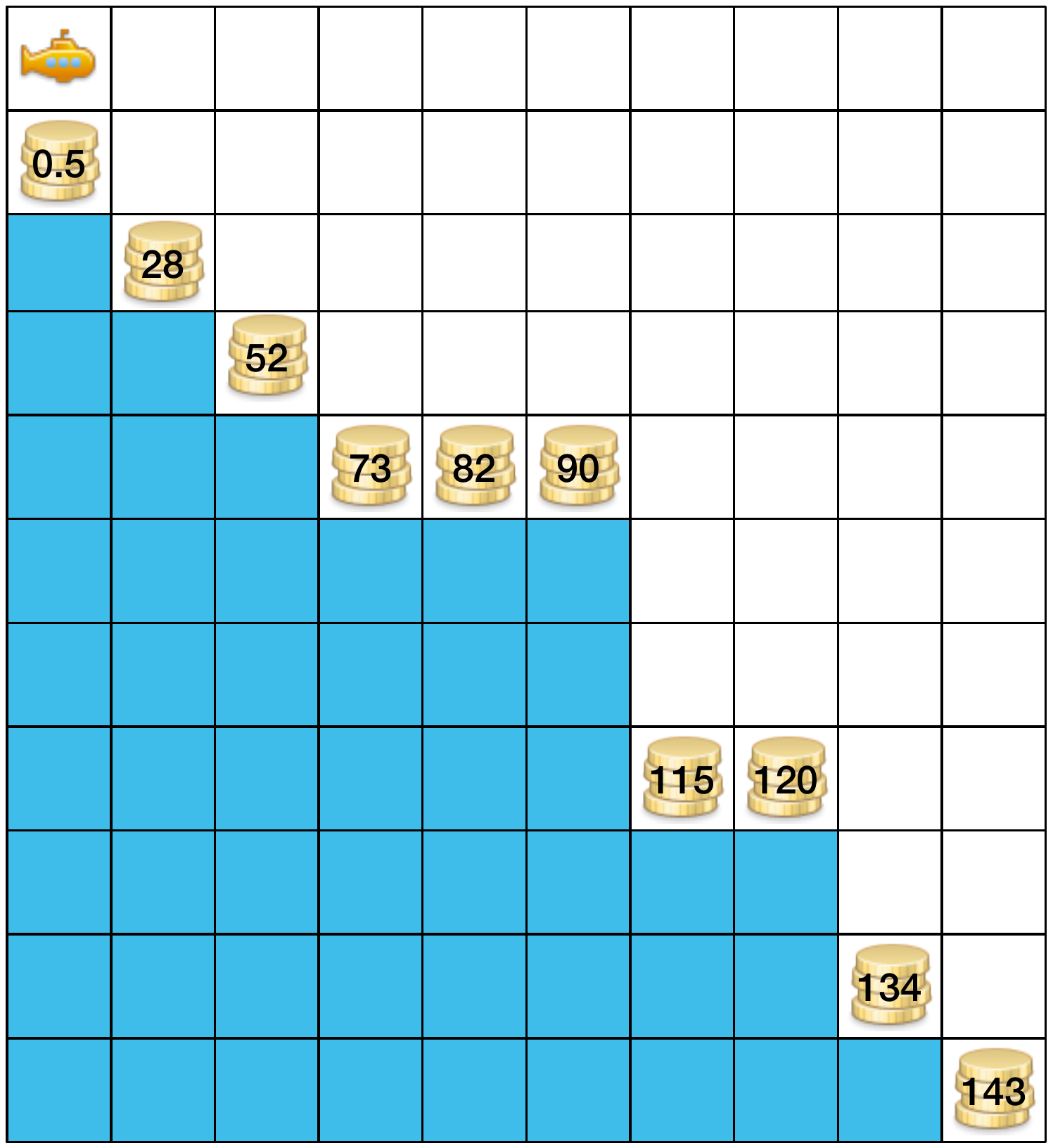}
	\vspace{-5mm}
    \caption{Image DST map.}
	\vspace{-6mm}
    \label{fig:deepsea-grid}
\end{wrapfigure}
To test the performance of our algorithms on a problem with a larger CCS, we adapt the well-known \emph{deep sea treasure (DST)}~\cite{vamplew2011empirical}  benchmark for MORL.
In DST, the agent controls a submarine searching for treasures in a $10 \times 11$ grid. The state $s_t$ consists of the current agent's coordinates $(x,y)$.
The grid contains 10 treasures that their rewards increase in proportion to the distance from the starting point $s_0=(0,0)$. The agent's action spaces is formed by navigation in four directions, and the map is depicted in \Cref{fig:deepsea-grid}.

At each time-step the agent gets rewarded for the two different objectives. The first is zero unless a treasure value was received, and the second is a time penalty of $-1$ for each time-step.
To be able to learn a CCS instead of a Pareto front, as it was in the original work~\cite{vamplew2011empirical}, 
 we have adapted the values of the treasures such that the value of the most efficient policy for reaching each treasure is in the CCS. The rewards for both objectives were normalised between $[0-1]$ to facilitate the learning.

\begin{wrapfigure}{r}{.33\textwidth}
	\vspace{-5mm}
    \includegraphics[width=1\linewidth]{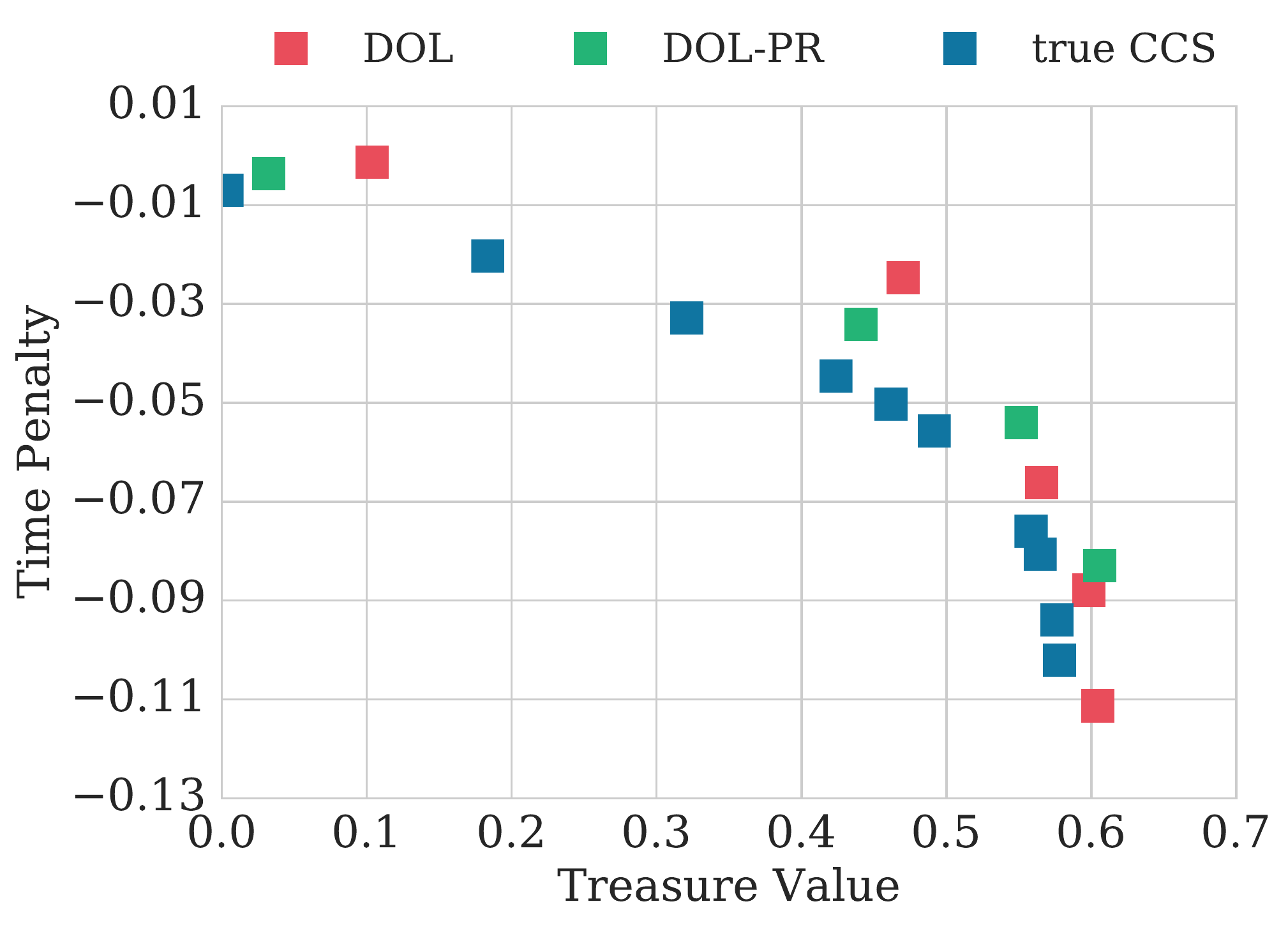}
	\vspace{-5mm}
    \caption{CCS after $4000$ episodes in DST raw version.}
    \label{fig:deepsea-map-dol-vs-dolpr-discovered-ccs}
	\vspace{-5mm}
\end{wrapfigure}

\begin{figure}[b!]
  \centering
  \begin{subfigure}[t]{0.33\textwidth}
    \includegraphics[width=1\linewidth]{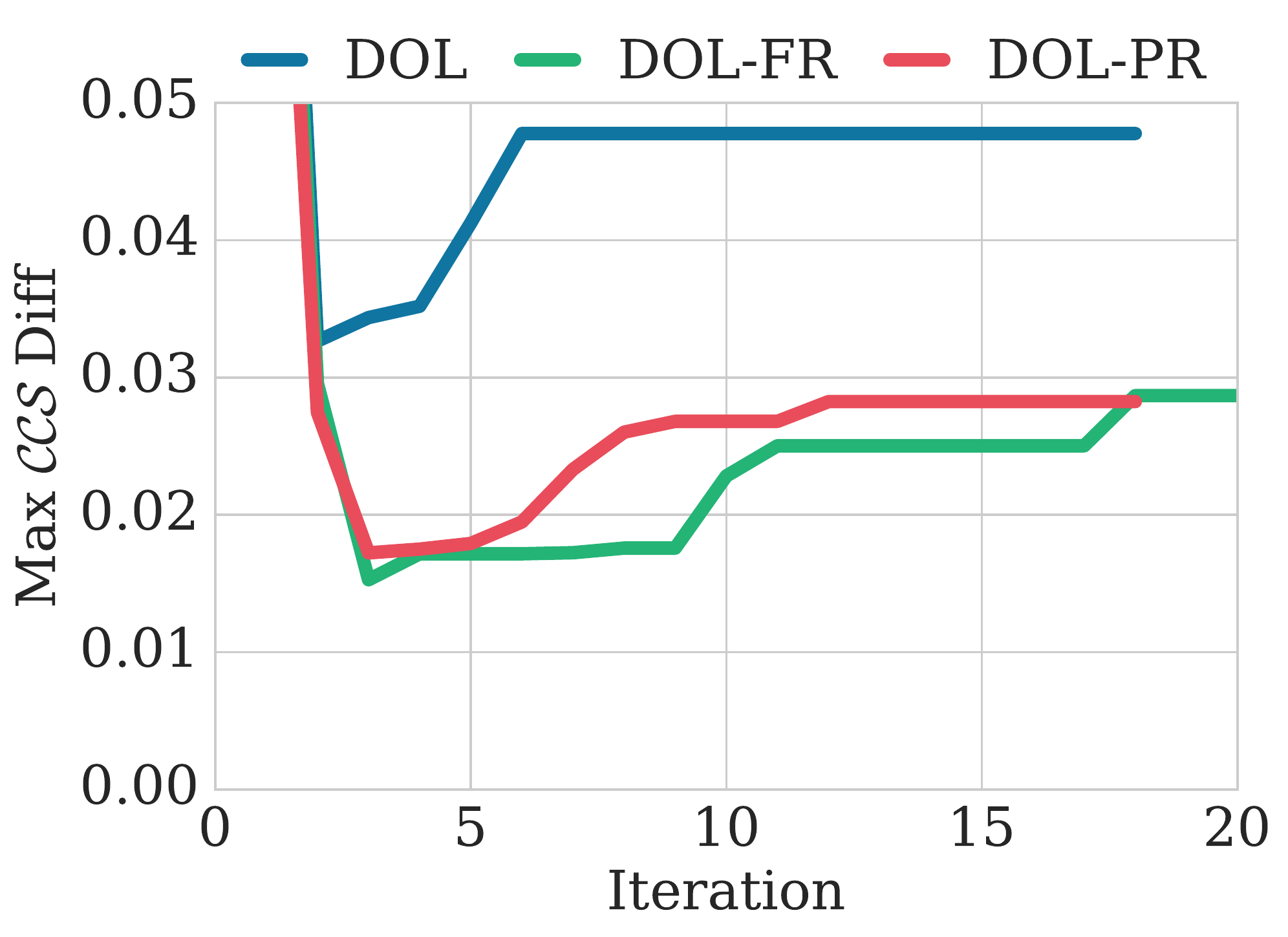}
    \caption{DST raw version.}
    \label{fig:deepsea-map-max-ccs-err}
  \end{subfigure}%
  \begin{subfigure}[t]{0.33\textwidth}
    \includegraphics[width=1\linewidth]{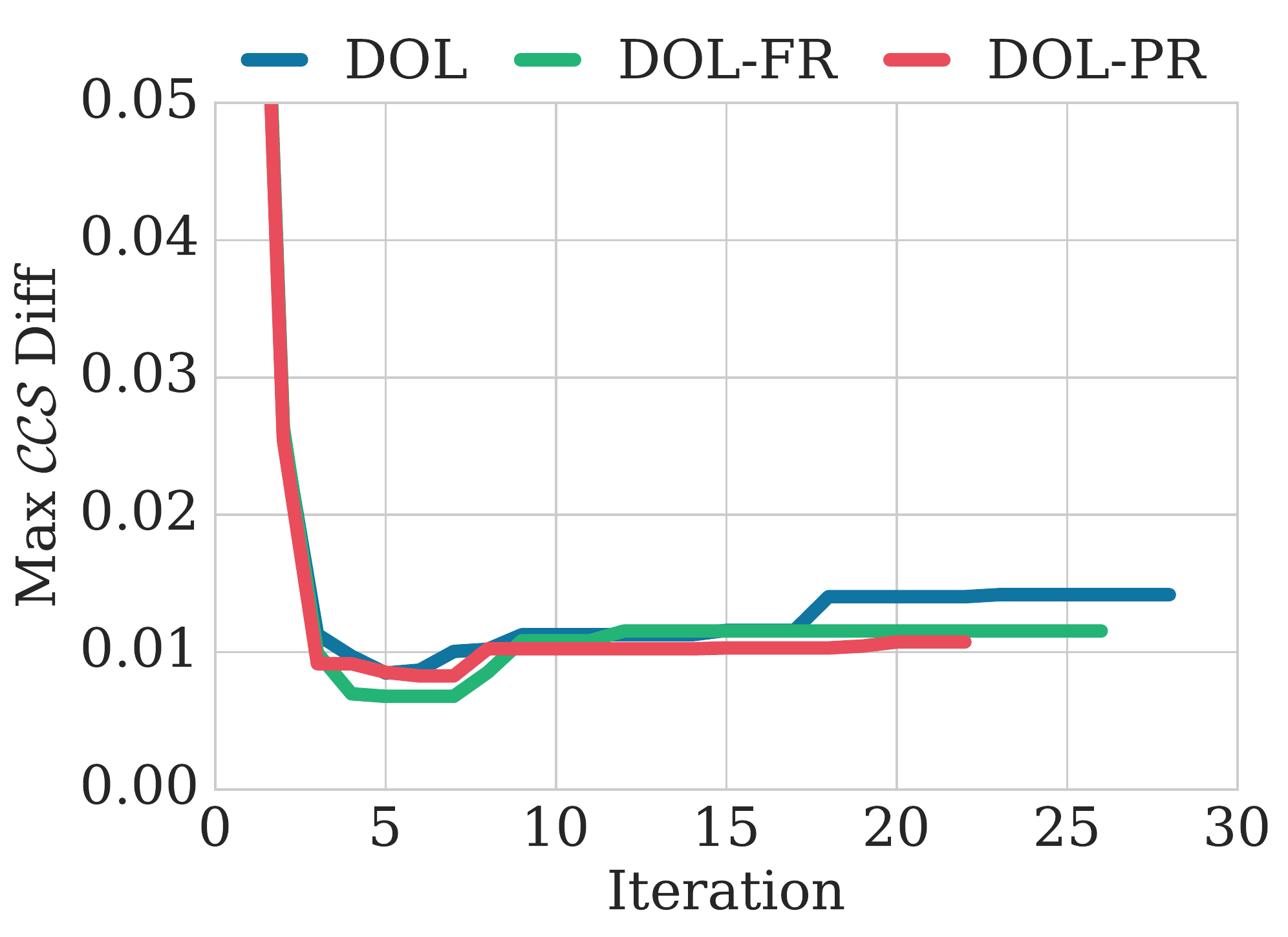}
    \caption{DST image version.}
    \label{fig:deepsea-image-max-ccs-err}
  \end{subfigure}%
  \begin{subfigure}[t]{0.33\textwidth}
    \includegraphics[width=1\linewidth]{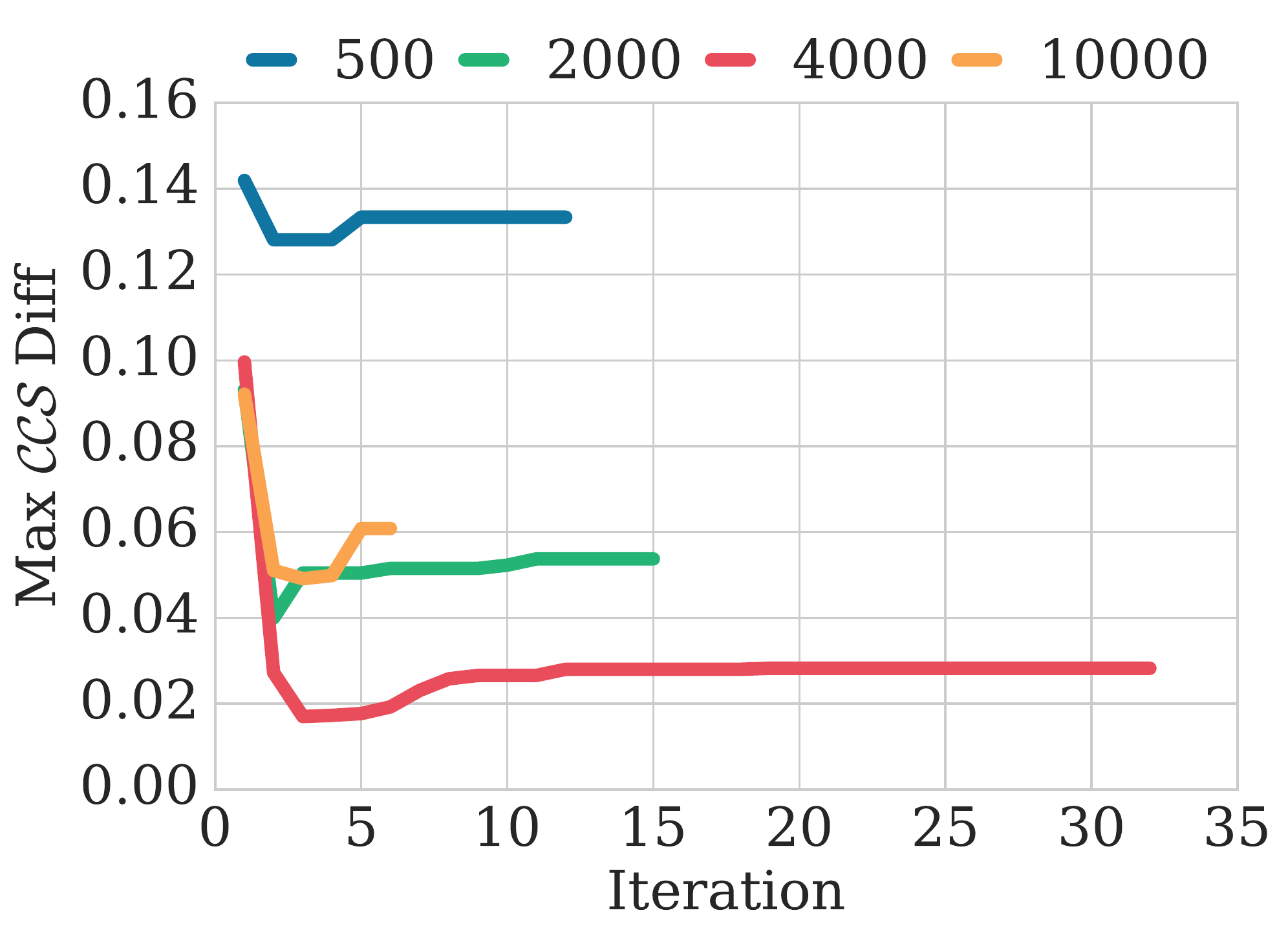}
    \caption{DST episodes vs accuracy.}
    \label{fig:deepsea-dolpr-episodes-vs-accuracy}
  \end{subfigure}  \caption{The Figures (a) and (b) illustrate the maximum CSS error in DST raw and image versions, respectivly. The results are averaged over $5$ experiments. Figure (c) shows the accuracy achieved for different number of episodes for DOL-PR.}
  \label{fig:deepsea-map-image-max-ccs-err}
\end{figure}

\textbf{Raw version.} We first evaluate our proposed methods, in a simple scenario, where the agent has direct access to the $s_t$. Hence, we employ a simple neural network architecture, to measure the maximum error in scalarised value with respect to the true CCS. The true CSS is obtained by planning with an exact algorithm on the underlying MOMDP. We refer to this error as \emph{Max CCS Error}. An analytical visualisation of measuring the true CSS and the discovered CSS difference, is illustrated in \Cref{fig:deepsea-map-dol-vs-dolpr-discovered-ccs}.
As it can be seen in \Cref{fig:deepsea-map-max-ccs-err}, DOL exhibits the highest error. Contrary to the preliminary expectations, having access to the raw state information $s_t$ does not make the feature extraction and reuse redundant. Furthermore, we discovered that when DOL-FR was used and the initialisation model already corresponded to an optimal policy, the miss-approximation error increased significantly, and less so for DOL-PR. We therefore conclude that our algorithms can efficiently approximate a CCS, and that reuse enables more accurate learning.

\textbf{Image version.} Similar to the raw version, our deep convolutional architectures for the image version, are still able to approximate the CCS with a high accuracy. As seen in \Cref{fig:deepsea-image-max-ccs-err}, the reuse methods show higher performance than DOL, and DOL-PR exhibits the highest stability as well.
This is attributed to the fact that the network has learned to encode the state-space, which paves the way towards efficient learning of the Q-values. DOL-PR exhibits the highest performance, as by resetting the last layer, we keep this encoded state-space, but we still allow DOL to train a new set of Q-values from scratch. We therefore conclude that DOL-PR is the preferred algorithm.

\textbf{Accuracy vs Episodes.} We further investigated the effects of the number of training episodes on the max CCS error. As can be seen in \Cref{fig:deepsea-dolpr-episodes-vs-accuracy}, the error is highly affected by the number of training episodes. Specifically, for a small number of episodes DOL-PR is unable to providine sufficient accuracy to build the CCS. It is interesting to note that though the error decreases up to $4000$ episodes, at $10000$ episodes the network is overfitting which results in lower performance.

\section{Related Work}
Multi-objective reinforcement learning~\cite{jair13,vamplew2011empirical} has recently seen a renewed interest. Most algorithms in the literature~\cite{BarrettCH,vanmoffaert14a,wiering2014model} are however based on an \emph{inner loop approach}, i.e., replacing the inner workings of single-objective solvers to work with sets of value vectors in the innermost workings of the algorithm. This is a fundamentally different approach, of which it is not clear how it could be applied to DQN, i.e., back-propagation cannot be transformed into a multi-objective algorithm in such a way. Other work does apply an outer loop approach but does not employ Deep RL~\cite{yahyaa2014scalarized,vanmoffaert2014novel,Natarajan05}. We argue that enabling deep RL is essential for scaling up to larger problems.

Another popular class of MORL algorithms are heuristic policy search methods that find a set of alternative policies. These are for example based on \emph{multi-objective evolutionary algorithms (MOEAs)}~\cite{coello2007evolutionary,handa2009multiobjective-eda-rl} or \emph{Pareto local search (PLS)}~\cite{kooijman15pareto}. Especially MOEAs are compatible with neural networks, but evolutionary optimisation of NNs is typically rather slow compared to back-propagation (which is what the deep Q-learning algorithm that we employ in this paper as a single-objective subroutine uses).

Outside of MORL, there are algorithms that are based on OLS but apply to different problem settings. Notably, the OLSAR algorithm~\cite{roijers2015point} does planning in multi-objective partially observable MDPs (POMDPs), and applies reuse to the alpha matrices that it makes use of to represent the multi-objective value function. Unlike in our work, however, these alpha matrices form a guaranteed lower bound on the value function and can be reused fully without affecting the necessary exploration for learning in later iterations. Furthermore, the \emph{variational OLS (VOLS)} algorithm~\cite{roijers2015variational}, applies OLS to multi-objective coordination graphs and reuses reparameterisations of these graphs that are returned by the single-objective variational inference methods that VOLS uses as a subroutine. These variational subroutines are not made OLS compliant, like the DQNs in this paper, but the value vectors are retrieved by a separate policy evaluation step (which would be suboptimal in the context of deep RL).

\section{Discussion} \label{sec:discussion}

In this work, we proposed three new algorithms that enable the usage of deep Q-learning for multi-objective reinforcement learning. Our algorithms build off the recent optimistic linear support framework, and as such tackle the problem by learning one policy and corresponding value vector per iteration. 
Further, we extend the main \emph{deep OLS learning (DOL)}, to take advantage of the nature of neural networks, and introduce full (DOL-FR) and partial (DOL-PR) parameter reuse, in between the iterations, to pave the way towards faster learning.

We showed empirically that in problems with large inputs, our algorithms can learn CCS with high accuracy. For these problems DOL-PR outperforms DOL and DOL-FR, indicating that a) reuse is useful, and b) doing partial reuse rather than full reuse effectively prevents the model from getting stuck in a policy that was optimal for a previous $\bf w$.
In future work, we are planning to incorporate early stopping technique, and optimise our model for the accuracy requirements of OLS, while lowering the number of episodes required.

\section*{Acknowledgements}
This work is in part supported by the TERESA project (EC-FP7 grant  \#611153).

\newpage
\setlength{\bibsep}{5pt}
\renewcommand*{\bibfont}{\small}
\bibliography{refs.bib,deeprl.bib}
\bibliographystyle{include/abbrvunsrtnat}

\end{document}